\documentclass{article}
\usepackage{graphicx} 
\usepackage{amsmath}
\usepackage{amsfonts}
\usepackage{graphicx}  
\usepackage[dvipsnames]{xcolor}
\title{Crack Detection in Infrastructure Using Transfer Learning, Spatial Attention, and Genetic Algorithm Optimization}
\author{Feng Ding }
\date{\today}

\begin{document}

\maketitle

\begin{abstract}
Crack detection plays a pivotal role in the maintenance and safety of infrastructure, including roads, bridges, and buildings, as timely identification of structural damage can prevent accidents and reduce costly repairs. Traditionally, manual inspection has been the norm, but it is labor-intensive, subjective, and hazardous. This paper introduces an advanced approach for crack detection in infrastructure using deep learning, leveraging transfer learning, spatial attention mechanisms, and genetic algorithm (GA) optimization. To address the challenge of the inaccessability of large amount of data, we employ ResNet50 as a pre-trained model, utilizing its strong feature extraction capabilities while reducing the need for extensive training datasets. We enhance the model with a spatial attention layer as well as a customized neural network which architecture was fine-tuned using GA. A comprehensive case study demonstrates the effectiveness of the proposed Attention-ResNet50-GA model, achieving a precision of 0.9967 and an F1-score of 0.9983, outperforming conventional methods. The results highlight the model’s ability to accurately detect cracks in various conditions, making it highly suitable for real-world applications where large annotated datasets are scarce.
\end{abstract}


\section*{Introduction}
Structural crack detection is vital for maintaining the safety and integrity of buildings, bridges, and other infrastructure. Cracks can compromise structural stability, cause accelerated wear, and lead to expensive repairs or catastrophic failures if not identified and addressed promptly \cite{crack1}. The traditional approach to crack detection typically involves manual inspections by trained professionals, which is labor-intensive, time-consuming, and prone to human error \cite{crack2}. Automated methods using image-based detection have emerged as promising alternatives, but they often rely on simpler machine learning algorithms that struggle with variability in materials, lighting, and surface conditions \cite{crack3}. Deep learning offers an appealing solution due to its ability to learn complex, hierarchical features from raw image data, leading to highly accurate and efficient crack detection.

However, one major challenge faced by deep learning approaches is the limited amount of labeled data specific to crack detection on different types of structures and materials. Unlike general-purpose image datasets, crack datasets are often small, difficult to acquire, and require expert annotation \cite{KATSIGIANNIS2023107105}. Training a deep learning model from scratch under these conditions is difficult due to the risk of overfitting and poor generalization. In this study, we employ transfer learning to overcome this limitation, using models pretrained on large datasets like ImageNet. These pretrained models, such as ResNet50, have already learned to extract meaningful features from millions of images \cite{transfer1}. By fine-tuning these models for crack detection, we can adapt their learned representations to new domains, leveraging their generalizability and reducing the need for extensive domain-specific data.

While transfer learning provides a strong foundation, it may not fully capture the unique characteristics of cracks, especially when working with different metal surfaces or structural conditions. Therefore, we enhance our model's performance by integrating a spatial attention mechanism. The spatial attention layer focuses on the most relevant regions of input images, such as edges and irregular patterns indicative of cracks \cite{la1}. This selective focusing reduces background noise and enhances the model's ability to localize cracks, particularly in challenging scenarios where cracks are subtle, partially obscured, or occur under difficult lighting conditions. 

The attention-enhanced output is further refined through a customized neural network that acts as an additional learning stage specifically designed to deepen and enhance the model's understanding of cracks. After the spatial attention mechanism highlights the most relevant regions of the image, the customized network processes these refined features to learn more intricate patterns. This network is tailored to capture complex relationships that might not be fully represented in the initial feature extraction layers or even by the attention mechanism alone.

To push the boundaries of our model's performance and introduce additional novelty, we employ a genetic algorithm (GA) to optimize the architecture of the customized neural network. Traditional methods for designing neural networks rely on manual tuning of hyperparameters, which can be inefficient and limited in scope \cite{elsken2019neuralarchitecturesearchsurvey}. By using a GA, we automate the search for optimal configurations of layers and neurons within the network. Each individual in the GA population represents a unique network structure, encoded as a chromosome specifying the number of layers and neurons. The GA evolves these individuals through selection, crossover, and mutation, guided by a fitness function based on model accuracy. This process explores a vast search space of possible architectures, allowing the model to adapt its complexity and depth to best suit the specific requirements of crack detection. By using a GA, we ensure that our model achieves superior accuracy and robustness, offering a scalable and adaptive solution to crack detection in various structural environments. 

This novel combination of transfer learning, attention mechanisms, and GA-driven optimization contributes to the growing field of crack detection by offering a more accurate, adaptive, and scalable solution to real-world structural monitoring challenges.

\section*{Literature Review}
Detecting cracks in construction has become a crucial area of focus for researchers in recent years\cite{CHEN2021102913}. The aim of crack detection is to identify and pinpoint cracks in structures and infrastructure, as these defects, if left untreated, can lead to structural failures. Research in this domain has concentrated on devising innovative, rapid, accurate, and non-destructive techniques for crack detection\cite{quick}. For example, Cha et al.\cite{cha1} explored the application of a Faster R-CNN model for detecting and identifying various structural damages, including cracks and corrosion. In another effort, Zhang et al.\cite{zhangetal} developed CrackNet, a deep convolutional neural network specifically designed for the semantic segmentation of cracks, enabling precise localization and classification.

Cha et al.\cite{cha2017} introduced a CNN-based technique for detecting cracks in concrete by employing object detection through bounding boxes, using data captured via handheld cameras, while Mei and Gul \cite{MEI2020119397} developed a generative adversarial network (GAN) and connectivity map to identify pavement cracks at the pixel level with images acquired from a GoPro camera. Both studies relied on datasets that requested human effort, which can be labor-intensive and challenging due to data acquisition complexities and variability in real-world conditions.

Meanwhile, open datasets have been made available to facilitate damage detection research. Dorafshan et al.\cite{dorafshan2018} compiled a dataset of 56,000 concrete crack images, categorized into crack and non-crack classes, while Xu et al.\cite{xuetal} developed a collection of 6,069 bridge crack images. However, despite these contributions, most studies in ML-based damage detection continue to depend on limited datasets, and public access to comprehensive datasets remains restricted \cite{KATSIGIANNIS2023107105}. 

Transfer learning has emerged as a powerful paradigm in the field of deep learning, aimed at mitigating the limitations posed by small or specialized datasets. In traditional machine learning, models are trained from scratch using domain-specific data, which often requires a vast amount of samples and significant computational resources. This approach becomes impractical in scenarios where data collection is costly, time-consuming, or constrained by the scarcity of samples. Transfer learning addresses this challenge by leveraging knowledge from models pre-trained on large, diverse datasets and adapting them to new, related tasks\cite{transfer1}. This process typically involves repurposing the feature extraction layers of a pre-trained network, such as ResNet, VGG, or MobileNet, and fine-tuning it with domain-specific data, thereby accelerating convergence and enhancing generalization capabilities. Zhu and Song\cite{zhusong} enhanced the VGG16 architecture using transfer learning to achieve precise classification of surface defects on concrete bridges. Similarly, Gopalakrishan et al. \cite{GOPALAKRISHNAN2017322} employed a pre-trained neural network to create a classifier capable of detecting cracks on both asphalt and concrete surfaces. In another approach, Zhang et al. \cite{kaige} proposed a framework that leveraged transfer learning with convolutional neural networks (CNNs) to categorize pavement images into "crack," "repaired crack," and "background" classes.

While transfer learning has proven to be a powerful strategy for leveraging pre-trained models and adapting them for crack detection tasks with limited data, it is not without its challenges. The pre-trained models may lack the capacity to focus on the context-specific details of cracks in various environments or structural materials, and the model performance has room for improvement \cite{BhalajiKharthik2024TransferLD}. This is where attention mechanisms offer a compelling solution. By guiding the model to emphasize the most relevant regions in input data, attention layers enhance the precision of the model’s predictions \cite{la1}. In the context of crack detection, spatial attention layers can prioritize and refine features in areas where cracks are more likely to occur, improving detection accuracy and robustness \cite{laziji}.

Attention mechanisms have been utilized to advance crack detection performance. Lan et al.\cite{attention1} introduced an enhanced YOLOv5 algorithm incorporating an attention mechanism to boost detection precision for cracks. Chen et al.\cite{lit7} developed an attention-based crack detection network (ACNet) designed to achieve more accurate localization of cracks from a visual perspective. Luo and Liu\cite{laziji} implemented an attention-driven SqueezeNet architecture combined with Gradient-weighted Class Activation Mapping (Grad-CAM) to automatically detect and visually interpret cracks in buildings.

While attention mechanisms have demonstrated their ability to improve crack detection by focusing on the most relevant regions of input images, further optimization of the model's architecture is often necessary to achieve the best performance. We propose to refine the attention-enhanced output through a customized neural network that acts as an additional learning stage specifically designed to deepen and enhance the model's understanding of cracks. 

Designing a customized neural network for optimal performance poses a challenge. Experiments on learning neural networks have shown that known methods of local and global optimization (gradient, stochastic, Newton, Hessian, etc.) require a significant number of learning steps, are sensitive to the accuracy of calculations, require a significant number of additional variables \cite{elsken2019neuralarchitecturesearchsurvey}.  Moreover, fine-tuning discrete hyperparameters, such as the number of layers in and the number of neurons in each layer, is a complex and resource-intensive process that can hinder model development \cite{numberlayer}. To address these challenges, Genetic Algorithms (GA) offer a compelling solution by providing a robust approach for model training and hyperparameter optimization, facilitating efficient exploration of the search space to achieve optimal configurations \cite{hyperga}, \cite{Lienkov2022DeepLO}.

In summary, while significant progress has been made in crack detection through the use of deep learning models, transfer learning, and attention mechanisms, challenges remain, particularly in optimizing model architecture and handling limited, often inaccessible datasets. The literature reveals that transfer learning has proven effective in mitigating data scarcity, attention mechanisms have enhanced detection precision by focusing on key regions of interest. Building upon these advancements, our approach further refines crack detection by incorporating a customized neural network designed to enhance the output from attention layers. To optimize the model architecture and overcome the complexities of hyperparameter tuning, we integrate a Genetic Algorithm (GA), enabling a more efficient search for optimal configurations.

\section*{Methodology}
\begin{figure} [!ht]
  \vspace{-1ex}
\begin{center}
\includegraphics[width=0.9\textwidth]{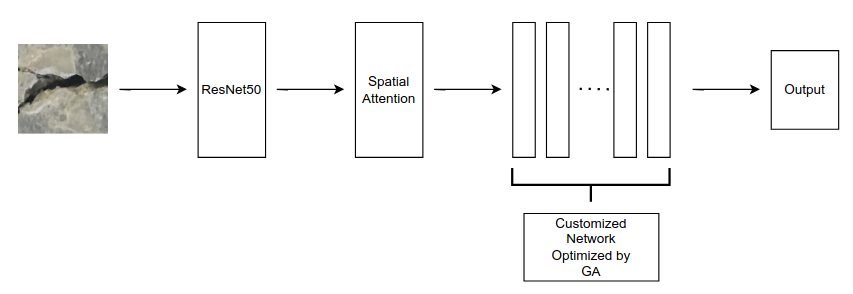}
\end{center}
  \vspace{-3ex}
    \caption{The flowchart of the proposed methodology}
   \label{fig:flowchart}
  \vspace{-1ex}
\end{figure}

In this study, we adopt a transfer learning approach that utilizes the ResNet50 architecture for feature extraction in conjunction with a spatial attention mechanism to improve crack detection accuracy. The motivation for using transfer learning stems from the limited availability of large, labeled datasets specifically for crack detection, which makes it impractical to train a deep neural network from scratch. The ResNet50 model, a 50-layer deep convolutional neural network known for its residual learning capabilities, is pretrained on the large ImageNet dataset. This pretrained model is used for its ability to extract rich, hierarchical features from input images without the need for extensive training on a crack detection dataset. We utilize the feature extraction layers from ResNet50, keeping the pretrained weights frozen and unaltered. By leveraging the generalizable features learned on a large, diverse dataset, the model benefits from robust initial representations that are highly transferable to the crack detection task, even with limited labeled data specific to structural cracks. Figure.\ref{fig:flowchart} illustrates the flow of the proposed method.

\subsection*{Spatial Attention Mechanism}

Once the features are extracted by ResNet50, they are passed through a spatial attention mechanism designed to refine the model’s focus on the most relevant regions of the image. This mechanism enhances the model's ability to prioritize areas likely to contain cracks, such as edges or irregular patterns, while reducing sensitivity to irrelevant background or noise. This targeted emphasis is particularly beneficial in challenging scenarios, such as detecting faint or partially obscured cracks in low-light conditions or amidst complex textures. By dynamically adjusting its focus, the spatial attention mechanism improves the model's ability to localize and identify subtle cracks, leading to higher precision and recall.

The spatial attention mechanism processes the input tensor $\mathbf{X} \in \mathbb{R}^{H \times W \times C}$, where $H$ and $W$ are the spatial dimensions, and $C$ is the number of channels. It computes an attention map that highlights significant spatial regions. The computation involves the following steps:

\begin{enumerate}
    \item \textbf{Average Pooling}: The average pooling operation calculates the mean value across all channels for each spatial position $(x, y)$, producing a map that captures the average channel response:
    \[
    \mathbf{A}_{\text{avg}}(x, y) = \frac{1}{C} \sum_{c=1}^{C} \mathbf{X}(x, y, c).
    \]
    This step provides a global perspective by emphasizing overall activity at each spatial location.

    \item \textbf{Max Pooling}: The max pooling operation computes the maximum value across all channels for each spatial position $(x, y)$, capturing the strongest channel response:
    \[
    \mathbf{A}_{\text{max}}(x, y) = \max_{c \in \{1, \ldots, C\}} \mathbf{X}(x, y, c).
    \]
    This step highlights the most dominant features, ensuring that even subtle yet crucial crack features are not overlooked.

    \item \textbf{Concatenation}: The results of the average pooling and max pooling operations are concatenated along a new channel dimension to form a combined representation:
    \[
    \mathbf{A}_{\text{concat}} = \text{Concat}(\mathbf{A}_{\text{avg}}, \mathbf{A}_{\text{max}}) \in \mathbb{R}^{H \times W \times 2}.
    \]
    This combined map provides complementary information, capturing both the general and the most prominent spatial patterns.

    \item \textbf{Convolution Operation}: A 2D convolution operation is applied to $\mathbf{A}_{\text{concat}}$ with a specified kernel size, followed by a sigmoid activation function to produce the attention map $\mathbf{M}$:
    \[
    \mathbf{M} = \sigma(\text{Conv2D}(\mathbf{A}_{\text{concat}})),
    \]
    where $\sigma$ ensures that the attention values are normalized between 0 and 1. This map reflects the importance of each spatial position, with higher values indicating regions more likely to contain cracks.

    \item \textbf{Attention Output}: The input tensor is element-wise multiplied by the attention map to produce the refined output tensor:
    \[
    \mathbf{X}'(x, y, c) = \mathbf{X}(x, y, c) \cdot \mathbf{M}(x, y),
    \]
    where $\mathbf{X}' \in \mathbb{R}^{H \times W \times C}$ represents the adjusted feature map. This step ensures that the model focuses on the most relevant regions while suppressing less important areas.
\end{enumerate}

Through this mechanism, the spatial attention layer enables the model to dynamically adapt its focus, enhancing its ability to detect cracks in complex and diverse settings. This refinement significantly improves the overall robustness and performance of the crack detection pipeline.

\subsection*{Optimization Using Genetic Algorithm}

The architecture further evolved by incorporating additional layers after the attention mechanism to increase the model's capacity for learning complex representations. The number and size of these added layers were determined through an optimization process guided by a genetic algorithm (GA). This approach allowed for exploration of a vast search space of possible network configurations, ensuring effective adaptation to the problem domain. The GA process we proposed is described as follows:

\subsubsection*{Genetic Algorithm Process}

1. \textbf{Initialization}: Create an initial population of \( N \) individuals, each representing a potential solution characterized by:
   \begin{itemize}
       \item Number of layers, \( L \in [1, L_{\text{max}}] \), where \( L_{\text{max}} \) is the maximum number of layers.
       \item Number of neurons per layer, \( \mathbf{n} = [n_1, n_2, \ldots, n_L] \) where \( n_i \in [n_{\text{min}}, n_{\text{max}}] \).
   \end{itemize}

2. \textbf{Fitness Function}: In the GA, the fitness of each individual was determined based on the model's classification accuracy on the validation set. The higher the validation accuracy, the higher the fitness score. The performance of each individual was evaluated using a fitness function \( f \) based on model accuracy:
   \[
   f(\text{individual}) = \text{validation accuracy}
   \]

3. \textbf{Selection}: During parent selection, individuals were chosen with a probability proportional to their fitness scores, ensuring that better-performing individuals had a higher chance of reproducing. This process also known as Roulette Wheel Selection is formally defined as follows:

3.1. \textbf{Fitness Calculation}: Let the fitness of an individual \( i \) be denoted by \( f_i \). The total fitness of the population of size \( N \) is given by:
   \[
   F_{\text{total}} = \sum_{i=1}^{N} f_i.
   \]

3.2. \textbf{Probability Assignment}: The probability \( p_i \) of selecting individual \( i \) is given by:
   \[
   p_i = \frac{f_i}{F_{\text{total}}}.
   \]
   This ensures that the probability of selection is directly proportional to the individual's fitness score.

3.3. \textbf{Cumulative Probability}: A cumulative probability distribution is constructed over the population. Let \( C_i \) denote the cumulative probability for individual \( i \), defined as:
   \[
   C_i = \sum_{j=1}^{i} p_j, \quad \text{where } 0 \leq C_i \leq 1 \text{ and } C_N = 1.
   \]

3.4. \textbf{Selection Process}: To select a parent, a random number \( r \in [0, 1] \) is generated. The individual \( i \) is chosen as a parent if:
   \[
   C_{i-1} < r \leq C_i,
   \]
   where \( C_0 = 0 \). This process is repeated to select the required number of parents.

The roulette wheel selection process ensures that individuals with higher fitness values are more likely to be selected for reproduction, but it also allows for exploration by giving lower-fitness individuals a chance to contribute to the next generation.

4. \textbf{Pairing}: After selecting parents based on their fitness values using roulette wheel selection, the paring (or mating) process for creating offspring is handled by random pairing. This method promotes diversity by ensuring a wide range of combinations, which can help explore a broader space of possible architectures. After selection, the parent pool is randomly shuffled to ensure there is no inherent ordering bias. The  parents are paired sequentially after shuffling. The process continues until all parents are paired. If there is an odd number of parents, the last individual is randomly paired with another from the pool to ensure all pairs are formed.

5. \textbf{Crossover}: Once the parents are selected, crossover is then applied to all the pairs to generate an offspring as follows:
Given two parents \( P_1 \) and \( P_2 \), represented as:
\[
P_1 = \{ L_1^{(1)}, L_2^{(1)}, \dots, L_{n_1}^{(1)} \}, \quad P_2 = \{ L_1^{(2)}, L_2^{(2)}, \dots, L_{n_2}^{(2)} \}
\]
where \( L_i^{(1)} \) and \( L_i^{(2)} \) denote the \( i \)-th layer of parents \( P_1 \) and \( P_2 \), respectively, and \( n_1 \) and \( n_2 \) are their respective layer counts, the process of creating a child \( C \) is defined as follows:

\textbf{Random Selection of the Number of Layers}: The number of layers \( n_c \) in the child is determined by:
\[
n_c = 
\begin{cases} 
n_1 & \text{with probability } 0.5, \\
n_2 & \text{with probability } 0.5.
\end{cases}
\]
This means the child inherits the total number of layers either from parent \( P_1 \) or from parent \( P_2 \).

\textbf{Random Selection of the Number of Neurons in Each Layer}: For each layer \( i \) in the child (where \( 1 \leq i \leq n_c \)), the number of neurons \( N_i^{(c)} \) in the child’s layer \( i \) is chosen randomly from the corresponding layers of the parents:
\[
N_i^{(c)} = 
\begin{cases} 
N_i^{(1)} & \text{if } i \leq \min(n_1, n_2) \text{ and with probability } 0.5, \\
N_i^{(2)} & \text{if } i \leq \min(n_1, n_2) \text{ and with probability } 0.5, \\
N_i^{(p)} & \text{if } i > \min(n_1, n_2), \text{ where } p \in \{1, 2\} \text{ is the parent with more layers}.
\end{cases}
\]
where \( N_i^{(1)} \) and \( N_i^{(2)} \) represent the number of neurons in the \( i \)-th layer of parents \( P_1 \) and \( P_2 \), respectively. If \( i > \min(n_1, n_2) \), the additional layers are inherited directly from the parent with the larger number of layers.

6. \textbf{Mutation}: After the crossover process, which creates offspring with potentially better configurations, mutation process is introduced to apply random changes to ensure genetic diversity and the exploration of novel solutions.
Given a new born offspring $B=\{L_1,L_2,…,L_n\}$, representing the layers $L_i$ of the neural network with $n$ layers, random mutations are applied to the number of layers and the number of neurons in each layer. The mutation for the number of layers is done with a probability \( p_{\text{add}} \) for adding a layer and \( p_{\text{remove}} \) for removing a layer. The mutation for the number of layers \( n_c \) in the child is defined as:

\[
n_c' = 
\begin{cases} 
n_c + 1 & \text{with probability } p_{\text{add}}, \\
n_c - 1 & \text{with probability } p_{\text{remove}}, \\
n_c & \text{with probability } 1 - p_{\text{add}} - p_{\text{remove}}.
\end{cases}
\]

where \( n_c' \) is the new number of layers in the child. Each layer of the child network undergos a mutation in its number of neurons with a probability \( p_{\text{neuron}} \). The number of neurons \( N_i \) in each layer \( L_i \) is mutated as follows:

\[
N_i' = 
\begin{cases} 
\text{random value from } [N_{\text{min}}, N_{\text{max}}] & \text{with probability } p_{\text{neuron}}, \\
N_i & \text{with probability } 1 - p_{\text{neuron}}.
\end{cases}
\]

where \( N_i' \) is the new number of neurons in layer \( L_i \).

7. \textbf{Termination}: Repeat the process for a specified number of generations or until convergence.

The GA was tasked with evolving the architecture by selecting the number of hidden layers and the number of neurons in each layer, with a maximum of 5 additional layers allowed to ensure computational feasibility. This approach ensured that the model could adapt effectively to the problem domain by balancing complexity and generalization.

\section*{Case Study}
\subsection*{Data Acquisitions and Pre-processing}
The proposed framework is tested on real-world images to detect cracks, demonstrating its effectiveness in recognition tasks. Our data sources include the Crack Forest Dataset (CFD), which consists of 329 images of urban concrete road cracks at an approximate resolution of 480 × 320 pixels \cite{shi2016automatic}, and the Concrete Crack Images for Classification Dataset (CCICD), featuring 20,000 images per class (crack and no-crack) at a resolution of 227 × 227 pixels in RGB format \cite{ozgenel2019}. Despite the availability of seemingly ample data, the CCICD images exhibit a high degree of homogeneity. This uniformity raises concerns about potential overfitting and inflated performance metrics during training and evaluation. To mitigate this issue, we selected a balanced subset of 6,000 images—3,000 per class—from various sources.

\begin{figure} [!ht]
  \vspace{-1ex}
\begin{center}
\includegraphics[width=0.9\textwidth]{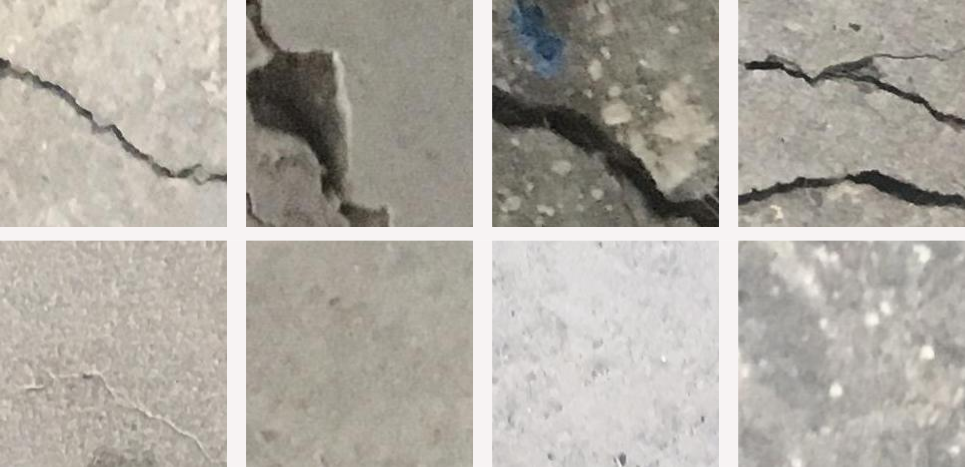}
\end{center}
  \vspace{-3ex}
    \caption{Demonstration of training set images}
   \label{fig:demo}
  \vspace{-1ex}
\end{figure}

Our final dataset comprises a diverse range of images, including clear, low-light, and obstructed examples, ensuring a more representative depiction of real-world crack conditions, as illustrated in Figure.\ref{fig:demo}. We categorized the images into two classes: "Negative" for images without cracks and "Positive" for images containing cracks. This mixed dataset blends actual field-collected data with existing databases, capturing diverse crack patterns such as linear, grid-like, and patch-like forms observed under different environmental conditions. By encompassing such variability, our dataset aims to enhance the model's adaptability to diverse real-world scenarios. To optimize computational efficiency, all images were resized to a standardized resolution of 224 × 224 pixels.

\subsection*{Study Design}
In this study, we utilize ResNet50 as our transfer learning base model. Previous work by Katsigiannis et al. \cite{KATSIGIANNIS2023107105} comprehensively evaluated various pre-trained models, including VGG16, VGG19, MobileNetV2, InceptionResNetV2, InceptionV3, and Xception. We specifically chose ResNet50 to explore its unique performance characteristics since it was not covered in their analysis. In our case study, we aim to illustrate the advantages of our proposed method. To achieve this, we compare the performance of the following models:
\begin{itemize}
    \item A CNN trained exclusively on our dataset.
    \item A baseline transfer learning model using ResNet50 connected directly to the output layer.
    \item A transfer learning model with ResNet50 enhanced by an attention layer.
    \item A transfer learning model with ResNet50, an attention layer, and a customized neural network further optimized through a genetic algorithm (GA).
\end{itemize}
Each successive model introduces an additional proposed feature on top of the previous one. By comparing their performance, we aim to demonstrate the incremental merits and effectiveness of each enhancement. We evaluate model performance using the metrics of Precision (P), Recall (R), and F1-score (F1), defined as: $P = \frac{TP}{TP + FP}$, $R = \frac{TP}{TP + FN}$ and $F1 = \frac{2PR}{P + R}$, where $TP$ stands for the number of true positives and $FN$ denotes the number of false negatives.

\subsection*{Result}

\begin{table}[ht]
\centering
\resizebox{\textwidth}{!}{
\begin{tabular}{|l|l|l|l|l|l|}
\hline
\textbf{Candidate Model} & \textbf{Size} & \textbf{No. Parameters} & \textbf{Precision} & \textbf{Recall} & \textbf{F1-score} \\ \hline
CNN                      & 98.44 MB      & 25.8M                   & 0.926              & 0.906            & 0.916            \\ \hline
ResNet50 (TL)            & 90.75 MB      & 23.78M                  & 0.9667            & 0.9797           & 0.973            \\ \hline
Att-ResNet50             & 90 MB         & 23.59M                  & 0.914             & 0.96             & 0.937            \\ \hline
Att-ResNet50-GA          & 91.69 MB      & 24.03M                  & 0.9967            & 1                & 0.9983           \\ \hline
\end{tabular}
}
\caption{Performance Comparison of Candidate Models}
\label{tab:model_comparison}
\end{table}

\begin{figure} [!ht]
  \vspace{-1ex}
\begin{center}
\includegraphics[width=0.9\textwidth]{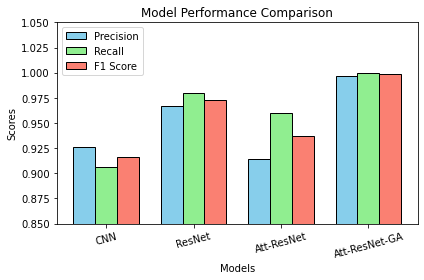}
\end{center}
  \vspace{-3ex}
    \caption{Barplot of model performance}
   \label{fig:compare}
  \vspace{-1ex}
\end{figure}

Table 1 presents a comparative analysis of various state-of-the-art models for crack detection, illustrating their respective performance across key metrics, while Figure \ref{fig:compare} offers a visual comparison. The convolutional neural network (CNN) trained on our dataset shows a commendable balance between precision and recall, highlighting its capability to accurately identify cracks while maintaining a moderate rate of false positives. Conversely, the ResNet50 model demonstrates its superior feature extraction abilities, as evidenced by its high recall score, minimizing missed crack instances and showcasing robust detection capabilities.

\begin{figure} [!ht]
  \vspace{-1ex}
\begin{center}
\includegraphics[width=0.9\textwidth]{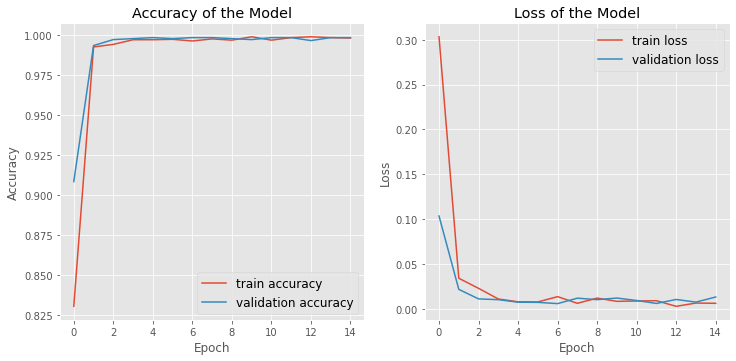}
\end{center}
  \vspace{-3ex}
    \caption{ The variation of loss and accuracy in the training and validation process.}
   \label{fig:train}
  \vspace{-1ex}
\end{figure}

The addition of a spatial attention layer to ResNet50 (denoted as Attention-ResNet50) yields  recall scores comparable to the original ResNet50, but its precision is much worse. This outcome suggests that incorporating a spatial attention layer alone does not guarantee significant performance improvement and may, in some cases, diminish the model’s efficacy. One possible explanation for this is the limited sample size available for training the attention layer parameters, which may restrict their optimization potential. However, this hypothesis is challenged by the performance of the Attention-ResNet50-GA model, which combines the attention mechanism with a customized neural network optimized using a genetic algorithm (GA). The GA algorithm identified $4$ layers for the customized neural network, with $66, 805, 218, 382$ being the number of the neurons in each layer, and despite having more parameters, the Attention-ResNet50-GA model achieves near-perfect results, with a precision score of 0.9967. Figure \ref{fig:train} illustrates the progression of loss and accuracy for both training and validation phases over the course of the model's training. During the initial epochs, the training loss decreases sharply, signifying the rapid adaptation of the model to the dataset. Simultaneously, training accuracy rises markedly, reflecting the model's ability to quickly capture fundamental patterns in the data. As training continues, the curves gradually stabilize, indicating that the model has achieved a balance between minimizing error and improving prediction accuracy. This stabilization suggests effective learning and an ability to generalize well to unseen data, as evidenced by the alignment of the validation metrics with those of the training set.

This exceptional performance underscores the effectiveness of our proposed enhancement, which leverages GA optimization to fine-tune the network and improve its focus on critical features for crack detection. By doing so, the model significantly reduces false positives while maintaining comprehensive true positive coverage. The resulting F1-Score of 0.9983 further emphasizes its balanced and superior detection capability. This makes the Attention-ResNet50-GA model particularly valuable for applications requiring high accuracy, efficiency, and reliability, positioning it as a leading approach in the field of crack detection.

\section*{Conclusion}

In this study, we proposed a hybrid approach for crack detection using deep learning, combining the power of transfer learning, spatial attention mechanisms, and genetic algorithm (GA) optimization to enhance model performance. The results demonstrated that our proposed method, Attention-ResNet50-GA, outperforms traditional models by achieving a significantly higher precision and recall, highlighting its ability to detect cracks with greater accuracy and minimal false positives. The GA-based optimization allowed the model to focus on the most relevant features, further improving its performance. This underscores the effectiveness of integrating these techniques for crack detection, particularly in scenarios requiring high accuracy and reliability.

While our approach has proven effective, it is not without its limitations. One major challenge is the computational expense involved in using the genetic algorithm to optimize the model. The GA requires a large number of iterations to explore the hyperparameter space, which can be very resource-intensive, especially when applied to deep learning models with complex architectures such as ResNet50. Despite leveraging transfer learning to reduce the time required for training, the addition of GA optimization increases the overall time and resource consumption significantly. This limitation could make our method less feasible in settings with constrained computational resources or time-sensitive tasks.

Moreover, the performance of the Attention-ResNet50 model demonstrated the importance of combining different techniques for optimal results. However, it was evident that simply adding the spatial attention layer to the pre-trained ResNet50 model did not yield significant improvements, suggesting that attention mechanisms require careful tuning and more data to realize their full potential. Future work could explore more efficient optimization techniques or hybrid models that combine the strengths of GA with other methods like reinforcement learning or evolutionary strategies. Furthermore, expanding the dataset to include a more diverse set of crack types and environmental conditions would help improve the robustness and generalizability of the model, making it suitable for a wider range of real-world applications.

\bibliography{sc_ref}
\bibliographystyle{nar}

\end{document}